\documentclass{article}
\usepackage{ijcai25}
\usepackage{xcolor}
\usepackage{times}
\usepackage{soul}
\usepackage{url}
\usepackage[utf8]{inputenc}
\usepackage[small]{caption}
\usepackage{graphicx}
\usepackage{amsmath}
\usepackage{amsthm}
\usepackage{booktabs}
\usepackage{algorithm}
\usepackage{algorithmic}
\usepackage[switch]{lineno}
\usepackage{subcaption}
\usepackage{cite}
\usepackage{multirow}

\title{Feature-Enhanced Machine Learning for All-Cause Mortality Prediction in Healthcare Data}

\author{
HyeYoung Lee$^{1,2}$
\and
Pavel Tsoi$^2$
\affiliations
$^1$The Department of Artificial Intelligence, Korea University\\
$^2$SPILab Corporation\\
\emails
uohesha@korea.ac.kr,
pauts@spilab.kr
}

\begin{document}

\maketitle

\begin{abstract}
Accurate patient mortality prediction enables effective risk stratification, leading to personalized treatment plans and improved patient outcomes. However, predicting mortality in healthcare remains a significant challenge, with existing studies often focusing on specific diseases or limited predictor sets. This study evaluates machine learning models for all-cause in-hospital mortality prediction using the MIMIC-III database, employing a comprehensive feature engineering approach. Guided by clinical expertise and literature, we extracted key features such as vital signs (e.g., heart rate, blood pressure), laboratory results (e.g., creatinine, glucose), and demographic information. The Random Forest model achieved the highest performance with an AUC of 0.94, significantly outperforming other machine learning and deep learning approaches. This demonstrates Random Forest's robustness in handling high-dimensional, noisy clinical data and its potential for developing effective clinical decision support tools. Our findings highlight the importance of careful feature engineering for accurate mortality prediction. We conclude by discussing implications for clinical adoption and propose future directions, including enhancing model robustness and tailoring prediction models for specific diseases.
\end{abstract}

\section{Introduction}
\label{introduction}
Accurate prediction of patient mortality is a critical challenge in healthcare, particularly within intensive care units (ICUs), where timely interventions can significantly impact patient outcomes \cite{kong2020using}. Early identification of high-risk patients allows for proactive resource allocation, personalized treatment strategies, and potentially improved survival rates. The increasing availability of electronic health records (EHRs) and large-scale clinical datasets, such as the Medical Information Mart for Intensive Care III (MIMIC-III) database \cite{mimic3}, has fueled interest in applying machine learning (ML) techniques to predict patient mortality and other critical outcomes. These rich datasets contain a wealth of longitudinal information, including patient demographics, vital signs, laboratory results, medications, diagnostic codes (ICD-9), and clinical notes, offering a comprehensive view of patient trajectories.
Predicting all-cause mortality is particularly important as it encompasses death from any cause, providing a holistic measure of patient risk. This is in contrast to cause-specific mortality, which focuses on death from a particular disease or condition. In the ICU setting, patients often present with multiple comorbidities and complex clinical presentations, making all-cause mortality a more relevant and practical endpoint for risk stratification.

While deep learning (DL) models have shown promise in various domains, including healthcare, traditional machine learning approaches, such as Random Forests (RF), often demonstrate competitive or even superior performance in clinical prediction tasks, especially when dealing with high-dimensional, noisy, and heterogeneous data characteristic of EHRs. RF models offer advantages such as robustness to outliers, implicit feature selection, and inherent interpretability, which are crucial for building trust and facilitating clinical adoption. Moreover, DL models often require substantial computational resources and large, meticulously curated datasets, which may not always be available in clinical settings.

This paper focuses on developing a robust and interpretable mortality prediction model for ICU patients using the MIMIC-III database. Our primary objective is to accurately identify patients at high risk of in-hospital mortality based on their clinical characteristics and interventions during their ICU stay. This predictive capability can empower healthcare professionals to make more informed and timely decisions, which could lead to better patient care and outcomes. Our approach employs a systematic pipeline: first, we perform rigorous data preprocessing and feature engineering, focusing on clinically relevant variables derived from the MIMIC-III dataset. Subsequently, we employ Least Absolute Shrinkage and Selection Operator (LASSO) regression for feature selection to identify the most influential predictors of mortality, mitigating the curse of dimensionality and enhancing model generalizability. We then train and optimize a Random Forest classifier using Grid Search to determine the optimal hyperparameters. Finally, we leverage SHAP values \cite{SHAP} to provide a comprehensive interpretation of the model's predictions, elucidating the contribution of each feature to the mortality risk assessment. This interpretability is crucial for gaining clinical trust and understanding the underlying factors driving mortality risk.

\begin{figure*}[htbp]
\centering
\includegraphics[width=\textwidth]{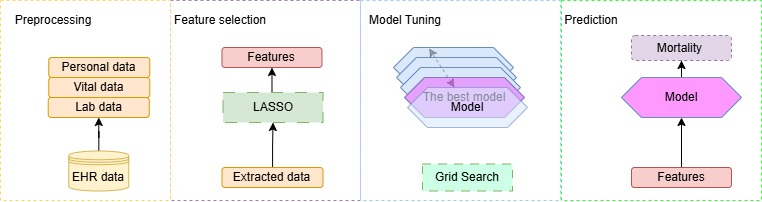}
\caption{Model architecture overview.}
\label{fig:architecture}
\end{figure*}

Our contributions are as follows:

\begin{itemize}
    \item Development of a robust and interpretable ML-based mortality prediction framework tailored for ICU patients.
    \item Demonstration of Random Forest's efficacy in managing complex, high-dimensional clinical datasets.
    \item Introduction of SHAP-based interpretability to enhance transparency and clinical trust in the model.
\end{itemize}

The remainder of this paper is organized as follows: Section \ref{sec:related_work} reviews related work in mortality prediction. Section \ref{sec:methodology} details the methodology, including data preprocessing, feature engineering, feature selection, and model training. Section \ref{sec:results} presents the experimental results. Finally, Section \ref{sec:conclusion} concludes the paper and discusses future research directions.

\section{Related Work}
\label{sec:related_work}

This section reviews related work on mortality prediction in healthcare.

\textbf{Methods Leveraging Relational Information:} Several studies have explored the use of relational information to improve healthcare prediction. \cite{2019ijcai} introduced Multimodal Attentional Neural Networks (MNN) to combine clinical notes, vital signs, and laboratory results, demonstrating the potential of integrating diverse data types. Similarly, \cite{2021ijcai} proposed a Cooperative Joint Attentive Network to handle irregular, multi-rate health data. These approaches, however, rely on multimodal data and complex neural architectures. While \cite{shang2021graphehr} proposed a Heterogeneous Graph Neural Network (HGNN) and \cite{liu2021kgdal} introduced a knowledge graph-based approach to capture relationships within EHRs, our work focuses specifically on structured tabular data, allowing us to investigate complex feature interactions within this data type using feature engineering techniques. Unlike these graph-based approaches, which require specialized graph representations and computational resources, our method utilizes a Random Forest model, prioritizing computational efficiency and explainability, which allows us to understand which features are influential in driving the model's output. Our feature engineering strategy, inspired by the concept of relational information explored in these studies, allows us to capture complex interactions within tabular data without the need for explicit graph representations.

\textbf{Mortality Prediction and Feature Engineering:} Recognizing the heterogeneity of ICU populations, several studies have focused on predicting mortality within specific patient subgroups. For instance, \cite{frontiers2021} developed machine learning models to predict mortality specifically for mechanically ventilated patients within the MIMIC-III database. Similarly, \cite{huang2025prediction} focused on predicting mortality based on short-term heart rate variability, while \cite{ding2021artificial} investigated mortality prediction in patients with pancreatitis. Additionally, \cite{li2022novel} proposed a novel approach for mortality prediction in patients with stroke. Other studies have also focused on specific patient populations and/or data handling techniques. \cite{kong2020using} explored mortality prediction using LASSO and grid search, but their analysis was limited to sepsis patients. Similarly, \cite{koshti2023hospital} applied SMOTE to address class imbalance, but their focus was exclusively on heart failure cases. These studies highlight the need for methods that can generalize across diverse patient populations and handle data complexities effectively. While these studies demonstrate the importance of considering patient-specific characteristics, they often focus on narrow patient populations. Our feature engineering strategy aims to capture these patient-specific nuances through the creation of interaction features, enabling our model to generalize across a more diverse ICU population represented in the MIMIC-III dataset. Unlike these subgroup-specific models, our approach aims for broader applicability while retaining the ability to capture relevant patient characteristics.

\textbf{Handling Irregular Data:} Addressing the challenges of irregular, incomplete, and noisy data is crucial in critical care. \cite{2023nn} proposed a neural network architecture tailored for more accurate in-hospital mortality prediction in the presence of missing data. Studies in \cite{nature2024}, \cite{sepsis1} utilized deep learning to predict sepsis-related mortality, addressing the dynamic nature of sepsis. While these studies address missing data through sophisticated modeling techniques, our work employs a simpler and more computationally efficient approach using imputation, as detailed in Section \ref{sec:methodology}. This allows us to focus on feature engineering and model interpretability while still effectively handling missing values.

\textbf{Traditional Machine Learning Approaches:} While deep learning has advanced the field, traditional methods remain competitive, especially with noisy, high-dimensional datasets. \cite{Li2021} used traditional machine learning methods such as XGBoost to predict in-hospital mortality in ICU patients with heart failure using the MIMIC-III database. Other studies have also explored traditional approaches; for instance, \cite{Liu2022} developed a nomogram combined with the SOFA score to predict in-hospital mortality for MIMIC-III patients. These studies demonstrate the effectiveness of traditional machine learning for mortality prediction. Building upon these approaches, this study applies a Random Forest model to the MIMIC-III dataset. However, unlike these previous works, which often rely on simple data augmentation techniques such as interpolation, we focus on a more sophisticated feature engineering approach, using LASSO for feature selection and grid search for hyperparameter optimization. This allows us to boost both accuracy and interpretability, providing clinically meaningful insights into the factors driving mortality prediction.

\section{Preprocessing and Data Description}
\label{sec:preprocessing}

\subsection{Dataset}

We used the Medical Information Mart for Intensive Care III database version 1.4 (MIMIC-III v1.4) for the study. MIMIC-III is a publicly available, single-center critical care database, approved by the Institutional Review Boards of Beth Israel Deaconess Medical Center (BIDMC, Boston, MA, USA) and the Massachusetts Institute of Technology (MIT, Cambridge, MA, USA). The database contains information on 46,520 patients who were admitted to various ICUs at BIDMC in Boston, Massachusetts, from 2001 to 2012 \cite{mimic3}. It includes charted events such as demographics, vital signs, laboratory tests, fluid balance, and vital status; records ICD-9 codes \cite{ICD-9}; and stores hourly physiological data from bedside monitors validated by ICU nurses. In addition, the database includes written evaluations of radiologic films by specialists for the corresponding time period. The use of the data, which consists of de-identified health information, has been deemed not human subjects research, and there was no requirement for individual patient consent \cite{mimic3}. 

\subsection{Clinical Characteristics}

The dataset includes vital signs (heart rate, respiratory rate, blood pressure, oxygen saturation, temperature) and laboratory measurements (serum creatinine, blood glucose, white blood cell count, lactate levels). Feature selection and engineering were guided by clinical knowledge and prior research. We aggregated time-series data into daily summary statistics (mean, maximum, minimum). Missing values were imputed using mean imputation for continuous variables and mode imputation for categorical variables. Patients with more than 50\% missing data in critical variables were excluded.

\begin{figure}[htbp]
\centering
\includegraphics[width=0.47\textwidth]{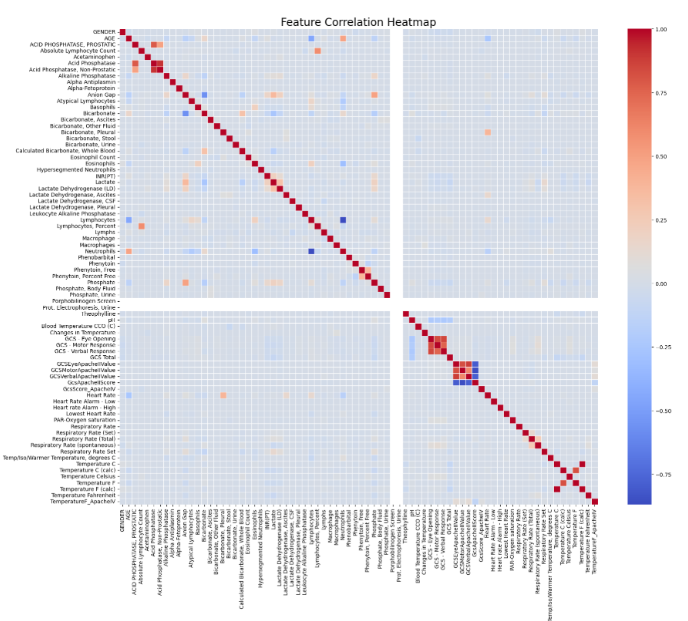}
\caption{Feature correlation matrix.}
\label{fig:feature_correlation}
\end{figure}

\section{Methodology}
\label{sec:methodology}
Our methodology involves data preprocessing, feature selection, model training, hyperparameter tuning, and interpretability analysis as presented in Figure \ref{fig:architecture}.

\subsection{Feature Selection and Preprocessing}

The first step is to preprocess the MIMIC-III \cite{mimic3} dataset by aggregating time-series data (e.g., vital signs) into summary statistics, such as daily means, maximums, and minimums. For a time series \( X_t \), where \( t \) represents time, we compute the following:

\begin{equation}
    \mu_t = \frac{1}{n} \sum_{i=1}^{n} X_{t_i}, \quad \max_t = \max_{i} (X_{t_i}), \quad \min_t = \min_{i} (X_{t_i}),
\end{equation}

Continuous variables are normalized using min-max scaling:

\begin{equation}
    X_{\text{norm}} = \frac{X - X_{\min}}{X_{\max} - X_{\min}},
\end{equation}

Categorical features are one-hot encoded to create binary vectors representing each category.

Feature selection is conducted in two stages. Initially, LASSO regression (Least Absolute Shrinkage and Selection Operator) is applied to shrink less relevant features to zero. The LASSO regression objective function is defined as:

\begin{equation}
    \mathcal{L}(\beta) = \sum_{i=1}^{n} \left( y_i - \mathbf{x}_i^T \beta \right)^2 + \lambda \sum_{j=1}^{p} |\beta_j|,
\end{equation}

where:
\begin{itemize}
    \item \( \beta \): the vector of regression coefficients (one for each feature),
    \item \( \lambda \): the regularization parameter that controls the sparsity of the model,
\end{itemize}

We further refine feature selection using Recursive Feature Elimination (RFE), which iteratively removes the least important features based on model performance, ensuring that the retained features are both clinically significant and statistically meaningful as presented in Figure \ref{fig:feature_correlation}.

\subsection{Data Augmentation}

Following feature selection, the dataset is split into training and testing sets using a 80:20 ratio. The training set \( D_{\text{train}} \) consists of 80\% of the data, while the testing set \( D_{\text{test}} \) consists of 20\%. The data splitting is done randomly to ensure that both sets represent the distribution of the data. 

Missing values are imputed using the mean for continuous variables and the mode for categorical variables. For a continuous variable \( X \), the imputation is defined as:

\begin{equation}
    X_{\text{imputed}} = \frac{1}{n} \sum_{i=1}^{n} X_i,
\end{equation}

For categorical variables, the most frequent category (mode) is used for imputation.

To handle class imbalance, where survivors significantly outnumber deceased patients, we apply the Synthetic Minority Over-sampling Technique (SMOTE) to generate synthetic samples for the minority class \cite{smote}. SMOTE creates synthetic samples by selecting a minority class instance and generating new samples along the line segments joining any/all of the \( k \)-nearest neighbors. The new samples are created as:

\begin{equation}
    X_{\text{new}} = X_{\text{sample}} + \lambda \cdot (X_{\text{neighbor}} - X_{\text{sample}}),
\end{equation}

where \( \lambda \) is a random value between 0 and 1 that controls the distance from the original sample. SMOTE is particularly useful in imbalanced datasets, as it helps balance the distribution of the classes by artificially generating new, plausible samples for the minority class. This technique aids the model in learning from a more balanced representation, thus improving its ability to classify minority class instances effectively.

\subsection{Model Training and Hyperparameter Tuning}

We utilize the Random Forest classifier due to its robustness in handling high-dimensional, noisy medical data. Random Forest is an ensemble learning method that combines multiple decision trees \( T_1, T_2, \dots, T_k \) to boost prediction accuracy and reduce overfitting. Each tree \( T_i \) is trained using a bootstrap sample \( D_i \) drawn with replacement from the training dataset. The final prediction \( \hat{y} \) is made by averaging the predictions of all trees (for regression) or by majority voting (for classification):

\begin{equation}
    \hat{y} = \text{mode} \left( T_1(x), T_2(x), \dots, T_k(x) \right),
\end{equation}

Hyperparameter tuning is performed using Grid Search with cross-validation to identify the optimal configuration, adjusting parameters such as the number of trees \( n_{\text{trees}} \), maximum depth \( d_{\text{max}} \), minimum samples per leaf \( s_{\text{leaf}} \), and the number of features considered at each split \( f_{\text{split}} \). The performance of the model for each configuration is evaluated using cross-validation, which minimizes overfitting and ensures generalizability. 

A \textit{Grid Search} is an exhaustive search method used for hyperparameter optimization. In this technique, we define a grid of hyperparameter values and systematically evaluate the model's performance for each possible combination. Specifically, we create a set of discrete values for each hyperparameter we want to tune. The Grid Search then explores every possible combination of these values, creating a "grid" of hyperparameter settings. This approach is typically combined with cross-validation, where the model is trained and tested on different subsets of the data to assess its robustness. The goal is to find the hyperparameter set that results in the best performance, typically measured using metrics like accuracy, precision, or recall, depending on the problem at hand. The combination of Grid Search and cross-validation ensures that we select hyperparameters that generalize well to unseen data and are not simply optimized for a specific training set.

In the context of Random Forest, Grid Search explores a predefined range of hyperparameters such as the number of trees \( n_{\text{trees}} \), the maximum depth \( d_{\text{max}} \), and other relevant settings. This method ensures that the best configuration is chosen to balance the model's ability to fit the data and generalize to new, unseen data, thus advancing the overall prediction accuracy and robustness. While Grid Search is effective at finding good hyperparameters, it can be computationally expensive, especially when the number of hyperparameters and the range of values for each hyperparameter are large. Other optimization methods, such as Randomized Search or Bayesian Optimization, can be more efficient in such cases.

\subsubsection{Model Interpretability with SHAP}

To upgrade the interpretability of the Random Forest model, we utilize SHapley Additive exPlanations (SHAP) values, which are grounded in game theory \cite{SHAP}. SHAP values provide a unified framework for interpreting model predictions by assigning each feature an importance value for a specific prediction.

\begin{figure}[htbp]
\centering
\includegraphics[width=0.5\textwidth]{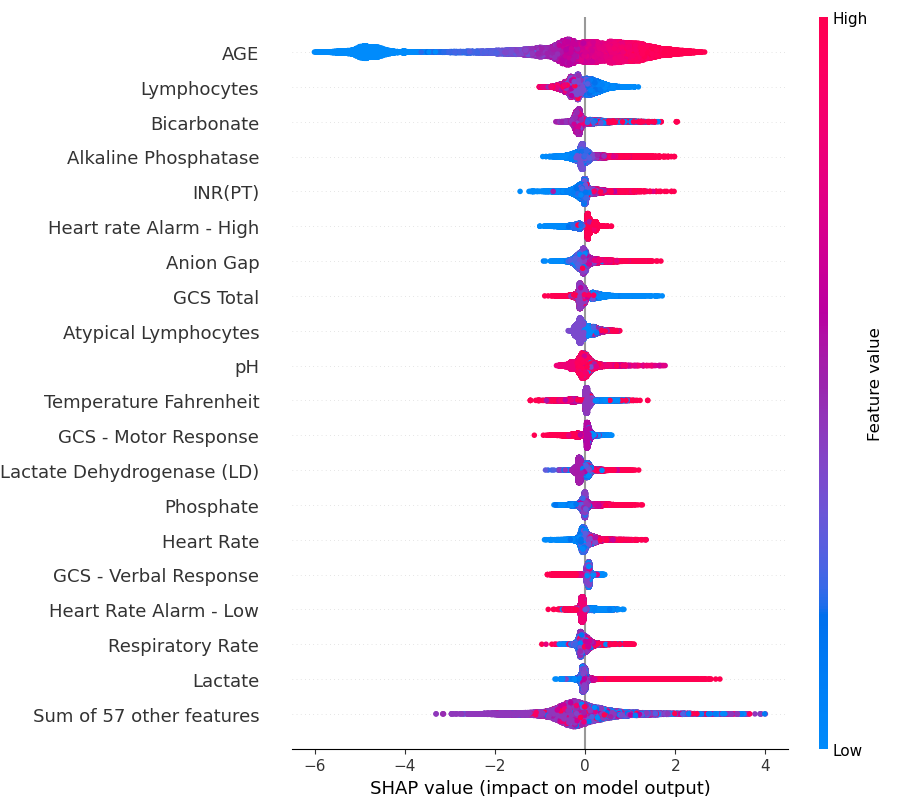}
\caption{SHAP values for feature importance.}
\label{fig:shap_value}
\end{figure}

For a given prediction \( f(x) \), where \( x \) is the input feature vector, the SHAP value for a feature \( j \), denoted as \( \phi_j(f) \), represents the average marginal contribution of that feature across all possible combinations of the other features. The formula for calculating SHAP values is:

\begin{equation}
\phi_j(f) = \sum_{S \subseteq N \setminus \{j\}} \frac{|S|!(|N|-|S|-1)!}{|N|!} \left[ f(S \cup \{j\}) - f(S) \right],
\end{equation}

where:
\begin{itemize}
    \item \( S \) is a subset of features excluding feature \( j \),
    \item \( N \) is the set of all features,
    \item \( |S| \) is the number of features in the subset \( S \),
    \item \( f(S \cup \{j\}) \) is the model's prediction when feature \( j \) is included in the subset \( S \),
    \item \( f(S) \) is the model's prediction when feature \( j \) is excluded from the subset \( S \).
\end{itemize}

In simpler terms, this formula calculates the difference in the model's output when feature \( j \) is included versus when it is excluded, considering all possible combinations of the remaining features. The weighted average of these differences yields the SHAP value for feature \( j \), quantifying how much that feature contributes to the change in the model's output. A positive SHAP value indicates an increased probability of mortality, while a negative SHAP value indicates a decreased probability of mortality. The magnitude of the SHAP value reflects the strength of the feature's influence.

SHAP values identify the key features driving mortality predictions, indicating whether their influence is positive or negative. Unlike traditional feature importance measures, which rank features based solely on their overall importance, SHAP values provide local explanations for individual predictions. This means we can understand how each feature contributed to a specific outcome, making SHAP particularly valuable in healthcare settings where understanding the factors influencing individual patient outcomes is crucial.

\begin{figure}[htbp]
\centering
\includegraphics[width=0.5\textwidth]{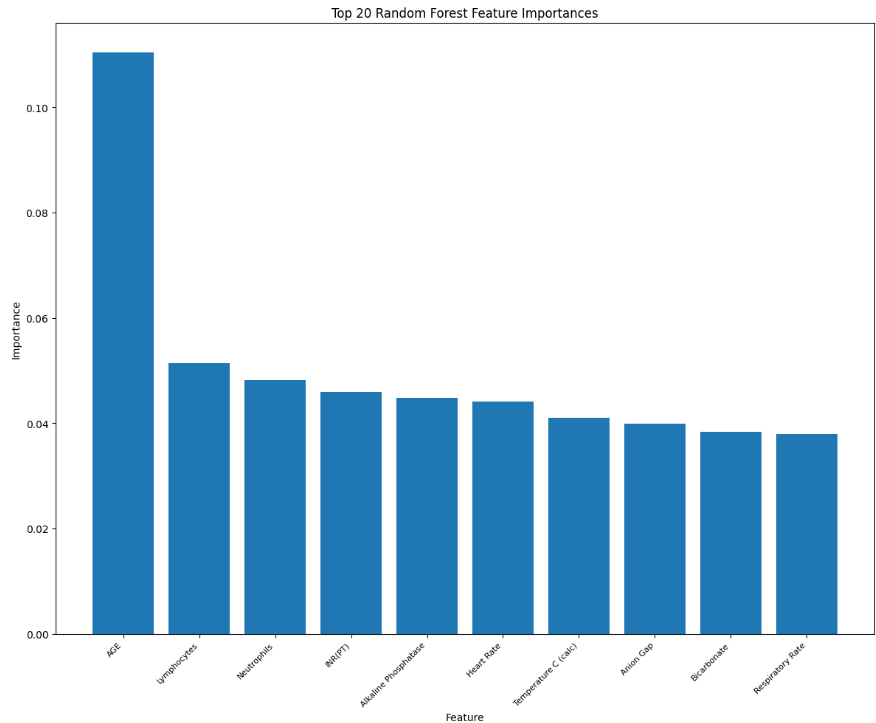}
\caption{Feature importance from the Random Forest model.}
\label{fig:feature_importance_RF}
\end{figure}

Furthermore, SHAP values can be aggregated to provide global explanations, offering insights into the overall importance of each feature across the entire dataset. This combination of local and global interpretability makes SHAP a powerful tool for understanding complex machine learning models in healthcare, enabling practitioners to make more informed decisions and potentially intervene more effectively in high-risk cases.

The use of SHAP values developes the transparency of the Random Forest model, aligning with the need for explainable AI in healthcare. By understanding which clinical variables have an impact in predicting patient mortality, healthcare professionals can improve their decision-making processes.

Figure \ref{fig:shap_value} provides a detailed analysis of feature importance using SHAP values for the Random Forest model. The SHAP summary plot shows the contribution of each feature to the model's predictions. Features with larger SHAP absolute values have a greater impact on the model's output. The color of the points indicates the feature value (red for high, blue for low), and the horizontal position shows the impact on the prediction. This visualization helps us determine the important clinical variables for predicting patient mortality and their associated positive or negative effects.

Figure \ref{fig:feature_importance_RF} displays the feature importance scores, derived directly from the Random Forest model. These scores reflect the relative contribution of each feature to the model's predictive accuracy, typically calculated based on how much each feature reduces impurity across all trees in the forest. A higher score indicates a stronger contribution to accurate predictions. This figure highlights the key clinical variables in predicting mortality, ranked from the age, followed by lymphocytes and neutrophils.

It is important to note the distinction between feature importance from the Random Forest and SHAP values. Random Forest feature importance measures the average decrease in impurity caused by a feature across all trees, providing a global measure of feature relevance. In contrast, SHAP values quantify the marginal contribution of a feature to individual predictions, offering both local-level and global insights. Although both methods provide valuable perspectives on the importance of features, they complement each other by offering different types of information. 

\begin{table}[htbp]
\caption{Model comparison on mortality prediction metrics.}
\resizebox{1\columnwidth}{!}{%
\begin{tabular}{lcccc}
\toprule
\textbf{Model} & \textbf{Precision} & \textbf{Recall} & \textbf{F1-score} & \textbf{Accuracy} \\
\midrule
RandomForest & \textbf{0.88} & \textbf{0.87} & \textbf{0.88} & \textbf{0.89} \\
XGBoost & 0.84 & 0.83 & 0.84 & 0.85 \\
TabNet & 0.77 & 0.79 & 0.78 & 0.78 \\
Anomaly Transformer & 0.76 & 0.75 & 0.75 & 0.77 \\
ResNet-18 & 0.66 & 0.66 & 0.66 & 0.67 \\
1dCNN & 0.31 & 0.50 & 0.38 & 0.62 \\
LSTM & 0.75 & 0.51 & 0.40 & 0.62 \\
MLSTM & 0.61 & 0.50 & 0.38 & 0.62 \\
GRU & 0.61 & 0.52 & 0.44 & 0.62 \\
GRU-D & 0.70 & 0.50 & 0.39 & 0.62 \\
\bottomrule
\end{tabular}}
\label{tab:model_comparison}
\end{table}

\section{Experiments}
\label{sec:results}
This section details the experimental setup and results obtained by evaluating the performance of our proposed model against a range of established baselines. We describe the datasets used, the experimental setup, the evaluation metrics, and provide a comprehensive analysis of the results.

\subsection{Experimental Setting}

\textbf{Baselines:}

\begin{figure*}[htbp]
\centering
\includegraphics[width=\textwidth]{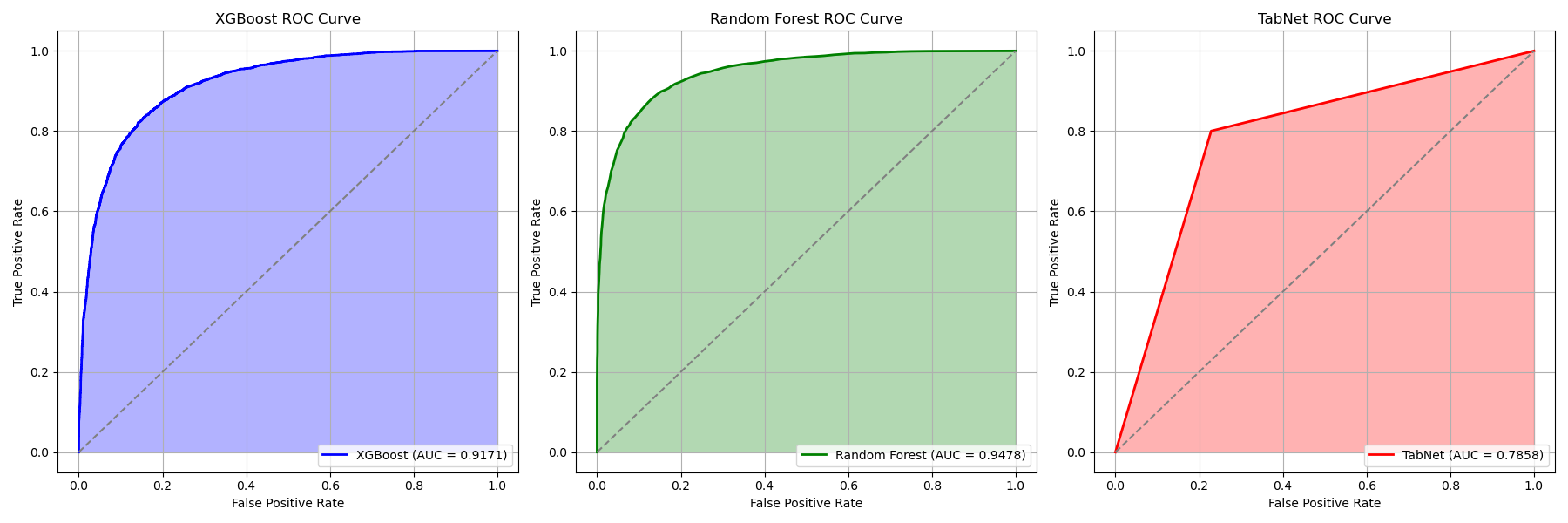}
\caption{ROC curves comparison for the XGBoost, Random Forest, and TabNet models.}
\label{fig:roc}
\end{figure*}

We compared our method against several established baselines, encompassing traditional machine learning models and deep learning architectures designed for time series analysis.

\begin{itemize}
    \item \textbf{Random Forest:} A robust ensemble learning method that constructs multiple decision trees during training and outputs the class that is the mode of the classes (classification) or mean/average prediction (regression) of the individual trees. Random Forests are known for their ability to handle high dimensionality and prevent overfitting. \cite{breiman2001random}
    \item \textbf{XGBoost:} (Extreme Gradient Boosting) Another powerful gradient boosting algorithm that builds an ensemble of decision trees sequentially, where each new tree corrects the errors of the previous ones. XGBoost is known for its speed and performance. \cite{chen2016xgboost}
    \item \textbf{TabNet:} A deep learning model specifically designed for tabular data. It uses sequential attention mechanisms to select relevant features at each decision step, providing interpretability and achieving high performance. \cite{arik2021tabnet}
    \item \textbf{1D CNN:} One-dimensional Convolutional Neural Networks apply convolutional filters along a single dimension (time in this case) to extract local features from the time series. They are effective in capturing temporal patterns.
    \item \textbf{LSTM:} (Long Short-Term Memory) A type of recurrent neural network (RNN) designed to address the vanishing gradient problem in traditional RNNs. LSTMs use gating mechanisms to regulate the flow of information, enabling them to learn long-term dependencies in sequential data. \cite{hochreiter1997long}
    \item \textbf{MLSTM:} (Multiplicative LSTM) An extension of the LSTM architecture that incorporates multiplicative gates, aiming to boost the model's ability to capture long-range dependencies. \cite{zhao2020do}
    \item \textbf{GRU:} (Gated Recurrent Unit) A simplified variant of the LSTM, with fewer gates, that still effectively captures temporal dependencies in sequential data. \cite{chung2014empirical}
    \item \textbf{GRU-D:} An adaptation of GRUs designed to handle time series with missing values. It incorporates information about the time intervals between observations into the model. \cite{che2018recurrent}
    \item \textbf{ResNet-18:} A deep residual network with 18 layers. ResNets use skip connections (residual connections) to address the vanishing gradient problem, enabling the training of very deep networks. While primarily designed for image recognition, they can be adapted for time series analysis by treating the time series as a 1D "image". \cite{he2016deep}
    \item \textbf{Anomaly Transformer:} A transformer-based model specifically designed for time series anomaly detection. It utilizes an association discrepancy to identify anomalies by comparing the associations between time series segments. \cite{xu2022anomaly}
\end{itemize}

We conducted experiments using an 80:20 train-test split with a fixed random seed of 42. All experiments were performed on a system running Windows 11, equipped with an NVIDIA GeForce RTX 3060 GPU with CUDA version 12.6, and an AMD Ryzen 7 6800H CPU. The Random Forest model, implemented using scikit-learn version 1.2.2, was trained with hyperparameters optimized via Grid Search and 5-fold stratified cross-validation. Although PyTorch version 2.5.1 was installed on the system, it was not utilized for the Random Forest model training, which relied solely on scikit-learn. The optimal hyperparameters found were: \texttt{n\_estimators=100}, \texttt{max\_depth=10}, \texttt{min\_samples\_split=2}.

\subsection{Evaluation Metrics}

We assess the performance of the Random Forest model using several classification metrics, including accuracy, precision, recall, F1-score, and the area under the receiver operating characteristic curve (AUC). Let \( TP \), \( TN \), \( FP \), and \( FN \) denote the number of true positives, true negatives, false positives, and false negatives, respectively. The metrics are defined as:

\begin{equation}
    \text{Accuracy} = \frac{TP + TN}{TP + TN + FP + FN},
\end{equation}

\begin{equation}
    \text{Precision} = \frac{TP}{TP + FP},
\end{equation}

\begin{equation}
    \text{Recall} = \frac{TP}{TP + FN},
\end{equation}

\begin{equation}
    \text{F1-score} = 2 \cdot \frac{\text{Precision} \cdot \text{Recall}}{\text{Precision} + \text{Recall}},
\end{equation}

\begin{equation}
    \text{AUC} = \int_0^1 \text{ROC}(t) \, dt,
\end{equation}

\subsection{Results}

As shown in Table \ref{tab:model_comparison}, the Random Forest model demonstrates superior performance across all metrics compared to the other evaluated models. It achieves a Precision of 0.88, Recall of 0.87, F1-score of 0.88, and Accuracy of 0.89. This indicates that the Random Forest model not only accurately identifies patients at high risk of mortality (high Precision) but also captures a large proportion of actual mortality cases (high Recall), resulting in a well-balanced performance (high F1-score) and overall high accuracy. The Receiver Operating Characteristic (ROC) curve, depicted in Figure \ref{fig:roc}, further corroborates the strong discriminatory power of the Random Forest model, showing a high Area Under the Curve (AUC), quantifying the model's ability to distinguish between positive and negative cases. A higher AUC value indicates better performance. Specifically, the Random Forest model achieved an AUC of 0.94, XGBoost achieved an AUC of 0.91, and TabNet achieved an AUC of 0.78.

The ROC curves in Figure \ref{fig:roc} visually compare the performance of the Random Forest, XGBoost, and TabNet models. The Random Forest model's curve is positioned higher and further to the left, indicating its superior ability to distinguish between positive and negative instances across various threshold settings.

In contrast to the strong performance of Random Forest, the deep learning models (TabNet, LSTM, MLSTM, GRU, GRU-D, ResNet-18, and Anomaly Transformer) generally underperform. While TabNet and Anomaly Transformer achieved relatively reasonable results, they still fall short of the Random Forest's performance. The recurrent neural networks (RNNs) and ResNet-18, in particular, exhibit significantly lower performance, especially in terms of Recall and F1-score. This suggests that these models struggle to capture the complex temporal dependencies in the MIMIC-III data or overfit to noise present in the dataset. The 1D CNN displays very poor performance, indicating it is not well-suited for this type of time-series data. XGBoost, while performing better than the deep learning models, still does not reach the level of Random Forest.

\subsection{Ablation study}

To further investigate the impact of different preprocessing and model training steps, we conducted an ablation study on the Random Forest model. This study involved systematically removing or modifying components of our methodology and evaluating the resulting performance. The results of the ablation study are presented in Table \ref{tab:ablation_study}.

\begin{table}[t]
\centering
\caption{Ablation study for each model.}
\label{tab:ablation_study}
\resizebox{1\columnwidth}{!}{%
\begin{tabular}{lcccc}
\toprule
\textbf{Module} & \textbf{Precision} & \textbf{Recall} & \textbf{F1-score} & \textbf{Accuracy} \\
\midrule
RandomForest & 0.87 & 0.87 & 0.86 & 0.87 \\
w/ SMOTE & 0.87 & 0.87 & 0.86 & 0.88 \\
w/ LASSO & \textbf{0.88} & 0.87 & 0.87 & 0.88 \\
w/ Grid Search & \textbf{0.88} & 0.87 & \textbf{0.88} & \textbf{0.89} \\
\bottomrule
\end{tabular}}
\end{table}

\section{Conclusion}
\label{sec:conclusion}

This study presents a comprehensive ML framework for predicting in-hospital mortality among ICU patients, leveraging robust feature engineering and the Random Forest algorithm. By combining clinical insights with machine learning, the proposed model achieves high accuracy and explainability from SHAP values, making it suitable for real-world clinical decision support systems. SHAP analysis further advances the transparency of the model, allowing healthcare providers to identify key risk factors and make informed decisions.

Future work will focus on several key areas to boost the model's capabilities and applicability. This includes validating the model across external datasets to ensure its generalizability and robustness in diverse clinical settings. Furthermore, we will explore the development of tailored prediction models for specific diseases from MIMIC-III dataset to provide more nuanced and accurate insights. These enhancements aim to extend the model's applicability across diverse clinical scenarios, ultimately contributing to improved patient outcomes and healthcare delivery.

\bibliographystyle{named}

\end{document}